\documentclass[10pt,twocolumn,letterpaper]{article}

\usepackage[pagenumbers]{cvpr} 

\usepackage[dvipsnames]{xcolor}

\definecolor{cvprblue}{rgb}{0.21,0.49,0.74}
\usepackage[pagebackref,breaklinks,colorlinks,citecolor=cvprblue]{hyperref}
\usepackage{gensymb}
\usepackage{amsmath}
\usepackage[capitalize]{cleveref}
\usepackage{graphicx}
\usepackage{xspace}
\usepackage{multirow}
\usepackage{makecell}

\usepackage[capitalize]{cleveref}
\crefname{section}{Sec.}{Secs.}
\Crefname{section}{Section}{Sections}
\Crefname{table}{Table}{Tables}
\crefname{table}{Tab.}{Tabs.}
\usepackage{url}            
\usepackage{amsfonts}       
\usepackage{nicefrac}       
\usepackage{microtype}      
\usepackage[dvipsnames]{xcolor}         

\newif\ifdrafting
\draftingtrue 
\ifdrafting
    \newcommand{\diff}[2]{{\color{red}\sout{#1}} {\color{green}#2}}
    \newcommand{\missing}[1]{{\color{orange}{#1}}}
    \newcommand{\vj}[1]{{\color{magenta}{\textbf{Varun: } #1}}}
    \newcommand{\ar}[1]{{\color{green}{\textbf{AR: } #1}}}
    \newcommand{\zi}[1]{\textcolor{magenta}{[Zubair: #1]}}
    \newcommand{\zk}[1]{\textcolor{blue}{[Zsolt: #1]}}
    \usepackage[textsize=tiny]{todonotes}
    \newcommand{\zkn}[1]{\todo[color=orange!20, size=\tiny]{Zsolt: #1}}
\else
    \newcommand{\diff}[1]{}
    \newcommand{\missing}[1]{}
    \newcommand{\vj}[1]{}
    \newcommand{\ar}[1]{}
    \newcommand{\zi}[1]{}
    \newcommand{\zk}[1]{}
    \newcommand{\zkn}[1]{}
\fi

\usepackage{algorithm}
\usepackage{algpseudocode}
\def\paperID{12828} 
\def\confName{CVPR}
\def\confYear{2024}

\usepackage[textsize=tiny]{todonotes}

\title{ICE-G: Image Conditional Editing of 3D Gaussian Splats\\}

\author{Vishnu Jaganathan\textsuperscript{1}, Hannah Hanyun Huang\textsuperscript{1}, Muhammad Zubair Irshad\textsuperscript{1,2},\\Varun Jampani\textsuperscript{3}, Amit Raj\textsuperscript{4}, Zsolt Kira\textsuperscript{1}\\ 
\textsuperscript{1}Georgia Institute of Technology, \textsuperscript{2}Toyota Research Institute, \textsuperscript{3}Stability AI, \textsuperscript{4}Google Research\\
{\tt\small \{vjaganathan3, hhuang474, mirshad7, zkira\}@gatech.edu,}\\
{\tt\small varunjampani@gmail.com, amitraj93.github.io}
}

\begin{document}
\twocolumn[{
\renewcommand\twocolumn[1][]{#1}%
\maketitle
\begin{center}
\captionsetup{type=figure}
\vspace{-0.20cm}
\includegraphics[width=0.95\textwidth]{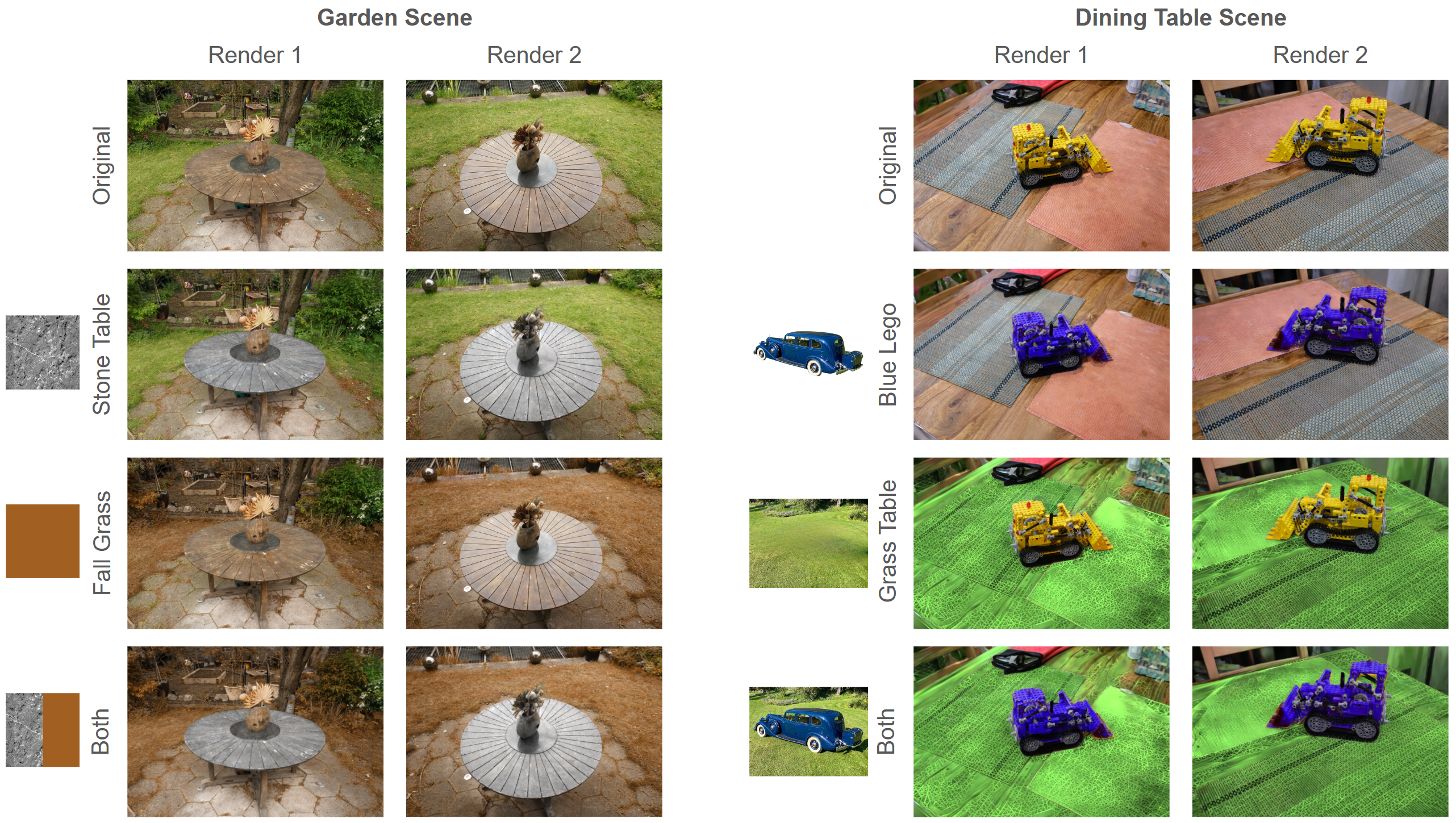}
\captionof{figure}{Our method, ICE-G, allows for quick color or texture edits to a 3D scene given a single style image, or mask selection on a single view. We show two rendered views of mask select editing for the Garden Scene where we apply stone texture to the table and fall colors to the grass (left). We also show two renders of correspondance based editing where we can transfer the color of the blue car to the lego and the texture of the grass to the table (right).
}
\label{fig:teaser}
\end{center}
}]
\begin{abstract}
\vspace{-0.45cm}
\looseness=-1
Recently many techniques have emerged to create high quality 3D assets and scenes. When it comes to editing of these objects, however, existing approaches are either slow, compromise on quality, or do not provide enough customization. We introduce a novel approach to quickly edit a 3D model from a single reference view. Our technique first segments the edit image, and then matches semantically corresponding regions across chosen segmented dataset views using DINO features. A color or texture change from a particular region of the edit image can then be applied to other views automatically in a semantically sensible manner. These edited views act as an updated dataset to further train and re-style the 3D scene. The end-result is therefore an edited 3D model. Our framework enables a wide variety of editing tasks such as manual local edits, correspondence based style transfer from any example image, and a combination of different styles from multiple example images. We use Gaussian Splats as our primary 3D representation due to their speed and ease of local editing, but our technique works for other methods such as NeRFs as well. We show through multiple examples that our method produces higher quality results while offering fine grained control of editing. Project page: \href{https://ice-gaussian.github.io/}{ice-gaussian.github.io}
\vspace{-20pt}
\end{abstract}    
\section{Introduction}
\label{sec:intro}

\looseness=-1 Editing of 3D scenes and models is an area of growing importance as applications like robotics simulation, video games, and virtual reality grow in popularity. Editable 3D representations can lead to dynamic and customizable environments in these applications, allowing artists, developers, and researchers alike to quickly iterate on projects and produce valuable content.

\looseness=-1 Recently, Gaussian Splats~\cite{kerbl3Dgaussians} have emerged as a powerful method to represent 3D objects and scenes, allowing for fast training and preservation of high-quality details. Prior to this, NeRFs (Neural Radiance Fields)~\cite{mildenhall2020nerf} have been used extensively to create scenes, and many techniques have been introduced to edit the color and texture of NeRFs. However, such editing has thus far been slow and limited in the types of edits possible.  Our work seeks to develop a general method that works on both Splats and NeRFs, supporting fast and high-quality style edits.

NeRF editing works can be categorized by their editing interfaces into text-based or image-based methods. Text-based approaches use text-image models for guidance, delivering results faithful to prompts but limited by the ambiguity of text descriptions for 3D scenes. This leads to uncertainties in conveying specific colors, styles, or textures, such as the exact shade of "light blue" or the precise pattern of a "sand texture."

Our image-based editing approach addresses the ambiguities of text-based methods, yet current techniques limit modifications to a single style image for the entire scene and lack the ability to transfer color or texture between different image parts or specify them per region. Additionally, 3D editing classifications based on changes—color, texture, shape, or their combinations—show that shape modifications often reduce image quality by converting 2D guidance into 3D using methods like Score Distillation Sampling (SDS)~\cite{poole2022dreamfusion} or Iterative Dataset Update (IDU)~\cite{instructnerf2023}, which generalize features at the expense of detail.

In this paper, we propose a method that aims to take the texture and/or color of different segmented regions from an editing image, and transfer them to corresponding segmented regions of a sampled set of 2D images from the original scene dataset in a 3D consistent manner. To generate high quality results, we restrict our method to editing color and texture while preserving shape. This editing image can either be a totally different object or an edited view from the original dataset. 

To do this, given a 3D model (e.g. Splat or NeRF), we propose to sample and edit a subset of the original data as a preprocessing step. Specifically, we use the Segment Anything Model (SAM)~\cite{kirillov2023segment} to find corresponding regions of both the editing image and the sampled views. For each region of each sampled view from the original dataset, we have to find the best corresponding region from the editing image to transfer style from. To find these matches, we utilize a custom heuristic which minimizes the distance between these mask regions in an extracted DINO~\cite{caron2021emerging} feature space. We then copy over colors by changing the hue and copy over textures by refitting them with Texture Reformer~\cite{wang2022texture}. To apply these updates onto the 3D model (e.g. Gaussian Splat), we then finetune it using L1 and SSIM losses for color, and a Nearest Neighbor Feature Matching (NNFM) loss~\cite{zhang2022arf} for texture.

We generate results for objects and scenes in the NeRF Synthetic~\cite{mildenhall2020nerf}, MipNeRF-360~\cite{barron2022mipnerf360}, and RefNeRF~\cite{verbin2022refnerf} datasets and compare against color and texture editing baselines to show qualitative improvement of our method. Overall, our main contributions are: 1) We provide a flexible and expressive mode of specifying edits, leveraging SAM and using a DINO-based heuristic to match image regions to the editing image in a multiview consistent manner, and 2) We provide fine grained control of choosing colors and textures for each part of the segmented editing view.

\section{Related Works}
\label{sec:related_works}

There are a few works that aim to edit 3D models, and they fall into a few categories. First are diffusion-based editing methods, which broadly try to lift inconsistent 2D image edits from text prompts into 3D via specialized losses. There are also local texture editing methods that are able to target regions and apply textures from a source image. Finally, since our method is capable of color editing as well, there are purely color edit methods we compare to that usually apply manually specified colors to images.

\subsection{2D Priors}
InstructNeRF2NeRF~\cite{instructnerf2023} adapts the InstructPix2pix~\cite{brooks2022instructpix2pix} 2D editing model to 3D, enabling edits in color, shape, and texture based on text prompts. Initially applied to select dataset images, these edits may lack multiview consistency but achieve 3D uniformity through Iterative Dataset Update (IDU), progressively refining more dataset examples. While this method supports extensive shape and color modifications, it tends to fall short in result quality and detailed texture rendering.

Vox-E~\cite{sella2023voxe} uses a voxel grid and diffusion model for updates, focusing on large feature edits. It processes views with text-guided noise predictions to align edits, but struggles with fine texture/color adjustments, often resulting in blocky textures or unintended area expansions.

Blended-NeRF~\cite{gordon2023blended} blends new objects or textures into scenes, guided by CLIP~\cite{radford2021learning} losses to match text inputs within a chosen 3D scene region. It modifies the scene's MLP with CLIP loss and blends colors and densities for the edits. While it achieves realistic textures, as a text-based method, it faces challenges in accurately conveying complex textures or specific regions without image input.

\subsection{Local Texture Editing}

S2RF~\cite{lahiri2023s2rf} introduces local texture editing for specific scene types, utilizing an object detection model along with SAM for precise region masking. This method applies NNFM loss from ARF for style/texture transfer onto masked areas, demonstrating the capability to apply varied textures to different scene parts.

Semantic-driven Image-based NeRF Editing (SINE)~\cite{bao2023sine} offers a method for 3D texture editing, leveraging a prior-guided editing field combined with original views. It uses a ViT~\cite{dosovitskiy2020vit} extracted style features to adjust textures, enabling localized edits. While it supports seamless rendering by merging template NeRF with the editing field, the process demands 12 hours of training per scene and faces compatibility issues with Gaussian Splats due to its unique rendering approach.

\subsection{Color Editing}
Decomposing NeRF for Editing via Feature Field Distillation~\cite{kobayashi2022distilledfeaturefields} allows color editing of NeRFs using text prompts. It generates a feature field for selecting and altering colors in 3D regions. Utilizing CLIP-LSeg~\cite{li2022languagedriven} and DINO~\cite{caron2021emerging} as 2D teacher networks, it learns an extra feature field integrated into the original NeRF, applying updates through photometric and feature loss functions. This approach enables soft 3D segmentation via a dot product between an encoded query and the feature field, facilitating text-specified color edits in 3D regions through modified rendering functions.

CLIP-NeRF~\cite{wang2021clip} learns a conditional NeRF representation that aims to separate appearance and shape information. CLIP embeddings are passed through appearance and shape mappers which extract the respective information and additively combine them with the conditional NeRF. These mapping layers are trained along with the NeRF via a CLIP similarity loss iterating over randomly sampled NeRF views. This method primarily edits color, but also shows minor shape changes on objects like cars and chairs.

RecolorNeRF~\cite{gong2023recolornerf} aims to decompose the scene into a set of pure-colored layers, and editing that pallet to change the color of the scene. This method achieves aesthetic results, but cannot distinguish between two different objects that have the same color in a scene. ProteusNeRF~\cite{wang2023proteusnerf} is able to rapidly edit the color of a NeRF by selecting a masked region and change its color, propogating the change into 3D. ICE-NeRF~\cite{Lee_2023_ICCV} finetunes the NeRF with the desired color edits, introducing techniques to preserve multiview consistency and avoid unwanted color changes.
\subsection{Concurrent Work in Gaussian Splat Editing}
Recently many methods have emerged that show editing capbilities on Gaussian Splats. One such example is GaussianEditor: Editing 3D Gaussians Delicately with Text Instructions~\cite{delicategaussianeditor}. This pipeline feeds the user prompt and scene description to an LLM to select regions of interest, and then applies a 2D diffusion prior to edit various views. Another similarly named paper GaussianEditor: Swift and Controllable 3D Editing with Gaussian Splatting~\cite{swiftgaussianeditor} introduces Hierarchical Gaussian Splatting, a technique to allow more fine grained editing via 2D diffusion priors. The user must select points on the screen, and change visual aspects manually in 2D for the edit to be carried over to 3D.

Instruct-GS2GSEditing: Editing 3D Gaussian Splatting Scenes with Instructions~\cite{igs2gs} is based off InstructNerf2Nerf and inherits the same advantages and disadvantages of that method. The authors use the same IDU method, but tune some hyperparameters to suit Gaussian Splatting. Another paper TIP-Editor: An Accurate 3D Editor Following Both Text-Prompts And Image-Prompts~\cite{tipeditor}  uses LoRA to personalize a diffusion model with the style of a reference image, and then uses this along with a user prompt to generate 2D edits. These edits are additionally bounded by a user specified region to contain edits.

Overall, these methods show some interesting results on editing using 2D Diffusion priors, but sometimes suffer the quality downgrade associated with diffusion models, and are not able to trasnfer style globally from a standalone 2D image.

\section{Method}
\begin{center}
    \centering
    \captionsetup{type=figure}
    \includegraphics[width=.5\textwidth]{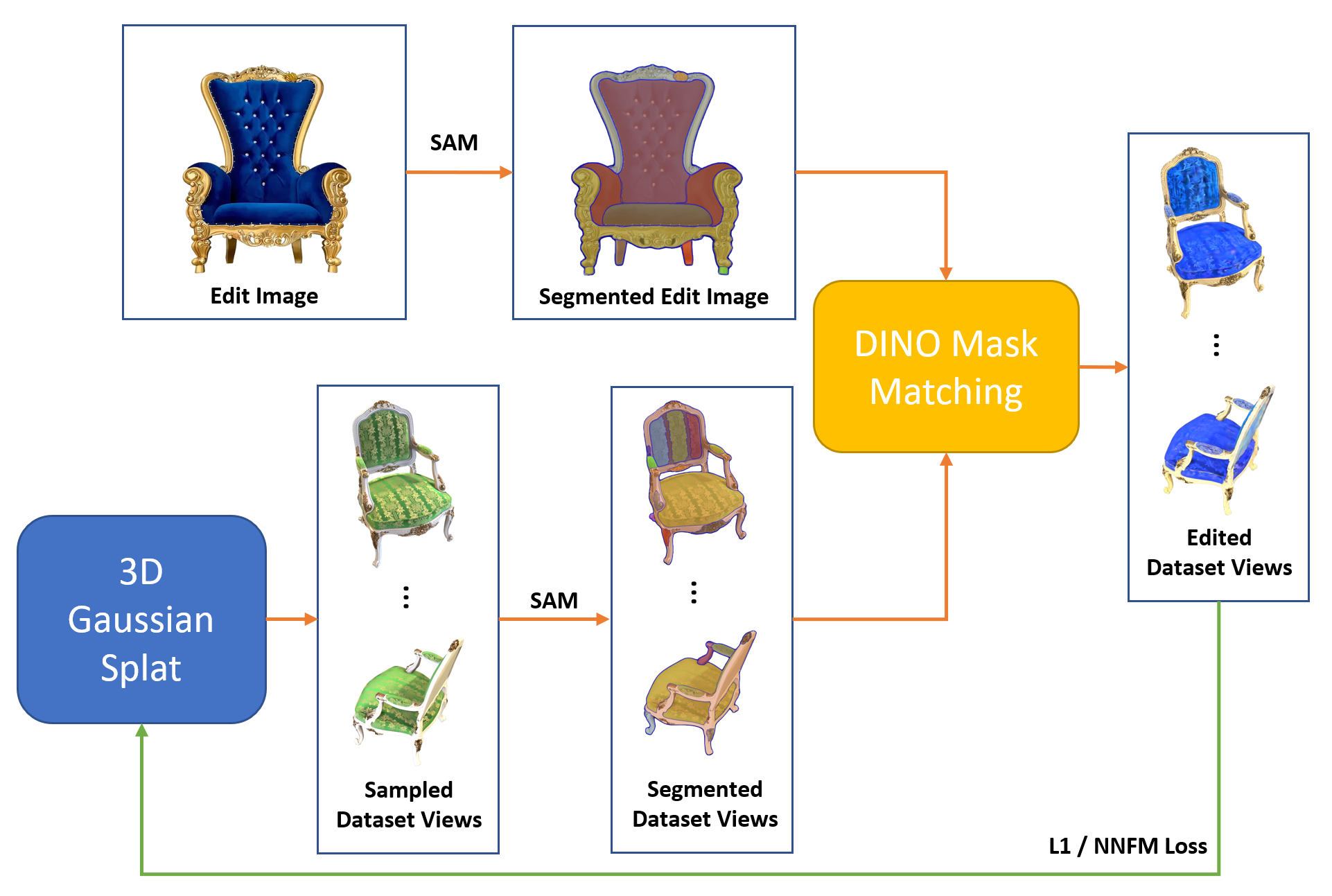}
    \captionof{figure}{The user supplied style image is segmented and its masked regions are matched with masked regions of sampled datset views via DINO correspondences. The color/texture is then transferred to those matching regions, and the splat is edited with this updated dataset.}
    \label{fig:training}
\end{center}

Our method supports different types of 3D models, and we primarily demonstrate it on top of Gaussian Splatting~\cite{kerbl3Dgaussians} due to its favorable speed. We also implement our method on a regular NeRF framework~\cite{lin2020nerfpytorch} for time comparison. There are two main interfaces, one for manual texture/color editing and another for automatically transferring these attributes from an example image as shown in Figure~\ref{fig:training}. The process differs only for creating the edit image, but uses the same segmentation, part matching, and texture/color losses across both. After making changes to the edit view, or choosing the conditional image, the algorithm is run on a number of sampled images where the style is transferred to these randomly sampled views. Color is naturally multiview consistent since only the hue is changed, and the underlying grayscale is preserved, so standard L1/SSIM loss is used to push the color updates. Since this is not the case for transferred texture updates, we employ the Nearest Neighbor Feature Matching (NNFM), originally proposed in ARF~\cite{zhang2022arf}, to make the texture change 3D consistent. Texture changes are done with this NNFM loss in a first round of iterations, and then color is changed with L1/SSIM losses in a second round, since we find that more vivid color is transferred via standard loss functions than NNFM.
\subsection{Preliminaries}
\subsubsection{Gaussian Splatting}
Gaussian Splats~\cite{kerbl3Dgaussians} is a recent 3D scene representation technique that allow for faster training and rendering. The scene is represented as a collection of 3D Gaussians, which are defined by position, covariance, opacity, and color. A given view is rendered from a differentiable rasterizer, which returns any given 2D view of this set of gaussians given standard NeRF-style viewing parameters. Since the rasterizer is differentiable, edits to the returned 2D image are backpropagated, and make the appropriate changes to the underlying gaussian representation. This method has the benefit of quick training time and producing more realistic textures than NeRFs in many cases. We base our method off of Gaussian Splats, but the technique works for NeRFs as well.
\subsubsection{SAM}
\looseness=-1 The Segment Anything Model (SAM)~\cite{kirillov2023segment} is a zero shot image segmentation model. It consists of a ViT encoder and mask decoder that produces the mask of each instance. This decoder is conditioned on either specific points, a box, or text to produce various masks. For the purpose of separating all parts of an object or scene, prompting with a grid of points is most effective. This produces distinct masked regions, which can be used as editing regions to apply new colors and textures.
\subsubsection{DINO}
Self-Distillation with No Labels (DINO)~\cite{caron2021emerging} is a self-supervised technique for training Vision Transformers (ViTs)~\cite{dosovitskiy2020vit} where a single ViT acts as both student and teacher. The student model learns from input data using standard methods, while the teacher model updates its weights through an exponential moving average of the student's weights, ensuring stable updates. This process encourages the student to learn generalizable and robust features by predicting the more stable teacher outputs. DINO facilitates effective ViT training without labeled data, leading to models that better focus on relevant image parts. The method's ability to identify pixel-wise image correspondences is further demonstrated in ~\cite{zhang2023tale}.
\subsubsection{Texture Reformer}
Texture Reformer~\cite{wang2022texture} introduces View-Specific Texture Reformation (VSTR) for transferring textures between image regions. By utilizing source and target semantic masks along with VGG feature extraction~\cite{Simonyan15}, it overlays textures from one area to another, adjusting to the new shape's contours. The technique employs patch grids and convolution for texture application, with statistical refinements ensuring realistic integration within the targeted masked regions.
\subsubsection{NNFM Loss}
Artistic Radiance Fields (ARF)~\cite{zhang2022arf} offers a method for infusing 3D NeRF scenes with style elements from 2D images. By processing 2D scene views alongside style images through a VGG-16 encoder, it applies a novel NNFM loss to match local features between the two, diverging from traditional Gram matrix losses that blend style details globally. This local matching technique ensures the preservation of texture specifics, marking a notable advancement over previous approaches.
\subsection{2D Editing}
\looseness=-1 There are two options for editing. Firstly, we can take a different conditional image(s), and copy styles from all parts onto the target objects views. In this approach, we use the texture-reformer module to bring all source textures onto a square array, so they can be cropped to the size of the target masks as necessary. We also store which colors correspond with which mask ids. Secondly, there is manual editing, where we start with any arbitrary view of the target object, and assign different styles to different regions. In both cases, we seek to generate a mapping of mask id to color/texture to find and copy these styles to the appropriate regions in the next steps. When editing, we can specify whether we want to copy only the color or the texture as well. This will determine whether we use the Texture Reformer module to extract textures, and whether we use NNFM loss or L1 loss alone in the downstream style applying steps, as opposed to sequentially.
\subsection{Segmentation}
The Segment Anything Model (SAM)~\cite{kirillov2023segment} is an encoder-decoder model that can be prompted with several grid points to make masks of most identifiable parts of an image. We use SAM to segment both the edit image and sampled views into their component parts, since it is the state of the art at this task and runs fairly quickly. Since we provide an option to manually specify which masks to edit, in our mask processing step, we allow users to specify a limit of N masks for simplicity. We choose the largest N-1 masks and group the rest of the image into the Nth mask. This basically enables the user to separate the editing view into any number of parts to have control over fine grained features. We first segment the editing image and store those masks, and segment each dataset view as we iterate over it.

\subsection{DINO Mask Matching}
We use DINO features to find which is the best editing image region to copy style from for each region of each of the sampled dataset views. Extracted DINO features have been shown to find corresponding pixels between two images ~\cite{caron2021emerging}. We use a similar feature extraction technique, but create a custom heuristic to measure the distance between two masks in the DINO feature space. First, we extract the DINO feature vector for each of the masks in the editing image, and store this information as it does not change. When iterating over a given dataset image, we extract DINO features after segmentation, and find the best matching region with the following heuristic:
\begin{equation}
M_P = \underset{E}{argmin} \frac{1}{N_E} \sum_{i \in P} (D(i) - D(E))^2
\end{equation}
To find the best match M for a given part P of a sampled dataset view, we find the editing image part E which is closest in the DINO feature space D.
\subsection{Texture Reformer}
We use the texture reformer~\cite{wang2022texture} module to copy textures from the editing image. Since we will be obtaining  masked regions (see next section), we can use that for our source semantic map, and the editing image itself for our source texture. For our target semantic mask, we can use the entire blank image of the same size. When applying textures to various different regions of the different dataset views, we can simply crop this full sized texture to shape. The reason we do this, rather than mapping the texture to each individual semantic mask for each view, is because we find empirically it does not matter, and time is saved by just doing this once. Running Texture Reformer per view does not lead to any more naturally view consistent results without NNFM  loss. This is done after the edit image is segmented in Figure~\ref{fig:training}.
\subsection{Applying Edits}
\subsubsection{Applying Color}
Applying a color change to a region of a view image is done in the HSV representation. In this representation the image is split up into the three channels of Hue, Saturation, and Value, rather than the standard RGB. The hue controls what color is expressed, the saturation controls how strong the color is, and the value controls how light or dark it is. The grayscale of an image, which contains the texture of the original view is the value. Therefore, to edit the color in a given target region, we copy over the average hue and saturation values of the source region, while leaving the value alone. If the user wants to brighten or darken a view overall, that can also be achieved by shifting the value field by a specified constant. 

Once this edit is made on the views, we use this as the data for training the edited 3D model. The loss function used is standard Structural Similarity Index (SSIM) and L1 interpolation used to train Gaussian Splatting:
\begin{equation}
\mathcal{L}_{GS} = \lambda \mathcal{L}_1 + (1 - \lambda)\mathcal{L}_{SSIM}
\end{equation}
\subsubsection{Applying Texture}
For texture, we either have manually specified a texture from a pattern image, or we automatically extracted texture from a matching region and expanded it as a pattern image with texture reformer. In either case we can crop this image sized texture to fit the mask region and add it. The texture will have the same pattern cropped to different viewpoints, and so is not 3D consistent. However, we again train a 3D model using this data and the NNFM loss, and over several iterations this will blend the image to be so. We find that using NNFM alone causes degradation in image quality and artifacts, and so we regularize it with the original Gaussian Splat training loss:
\begin{equation}
\mathcal{L}_{texture} = \mathcal{L}_{NNFM} + \alpha \mathcal{L}_{GS}
\end{equation}
This texture transferring often imprints the correct pattern on the Gaussian Splat, but leads to color appearing washed out. Thus, we follow up with some iterations of the color applying stage, copying over the average hue and saturation of the texture image.
\section{Results}
\subsection{Experiment Details}
We discovered in our ablation study in Figure 3, that sampling around 20\% of the images in the dataset for editing is sufficient for a good quality result. The color editing stage is run for around 2000 iterations, and the texture editing takes 3000 iterations to fully stylize the Gaussian Splat like the editing inputs. For the L1+SSIM portion of the loss, we use the Gaussian Splat implementation default interpolation. For texture loss, we find that adding 50\% of the original loss as a regularizer to the NNFM loss works best.
\begin{center}
    \centering
    \captionsetup{type=figure}
    \includegraphics[width=.5\textwidth]{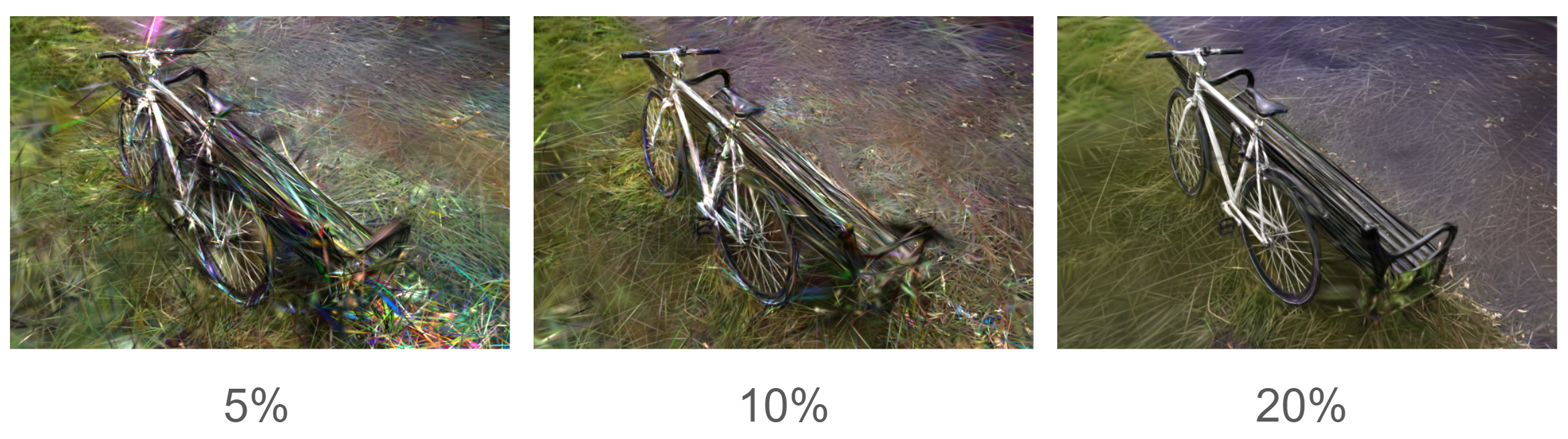}
    \captionof{figure}{Comparing different dataset sampling rates for turning the road to a river. Sampling 5\% or 10\% of images from a dataset to edit results in numerous artifacts and other degradations, and quality peaks at around 20\% sampling.}
    \label{fig:ablation}
\end{center}
\begin{center}
    \centering
    \captionsetup{type=figure}
    \includegraphics[width=.45\textwidth]{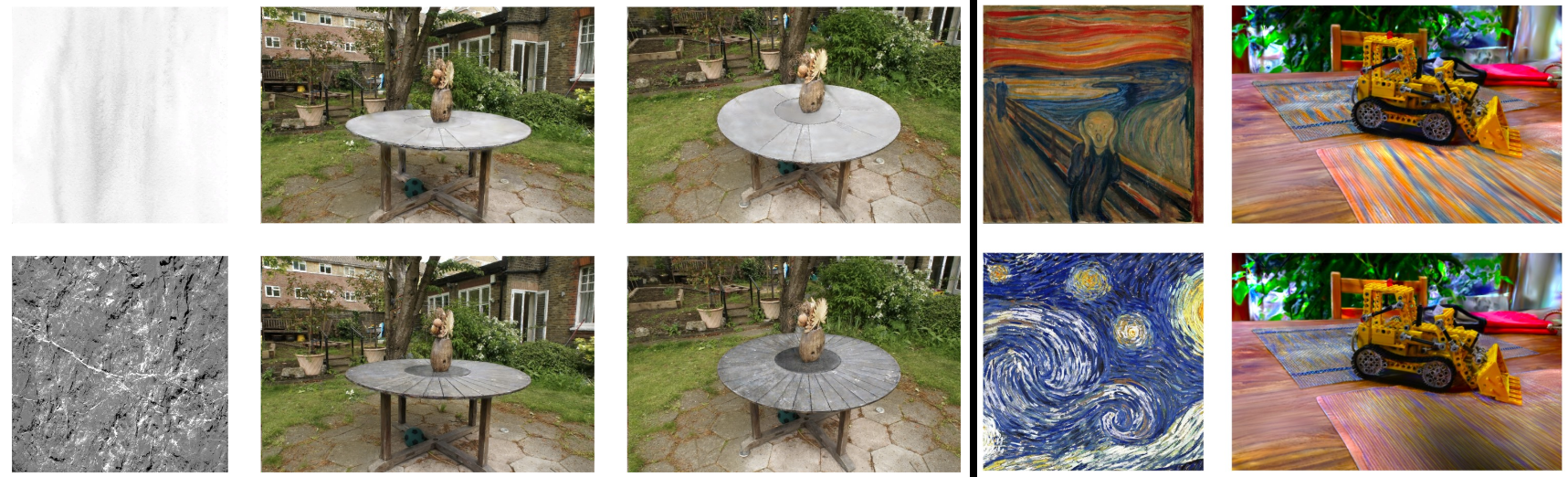}
    \captionof{figure}{Adding texture to the garden table from mip-NeRF360 (left). Using paintings to texture the table (right).}
    \label{fig:multi_garden}
\end{center}
\subsection{Texture Editing}
\label{sec:tex_editing_results}
\begin{center}
    \centering
    \captionsetup{type=figure}
    \includegraphics[width=.45\textwidth]{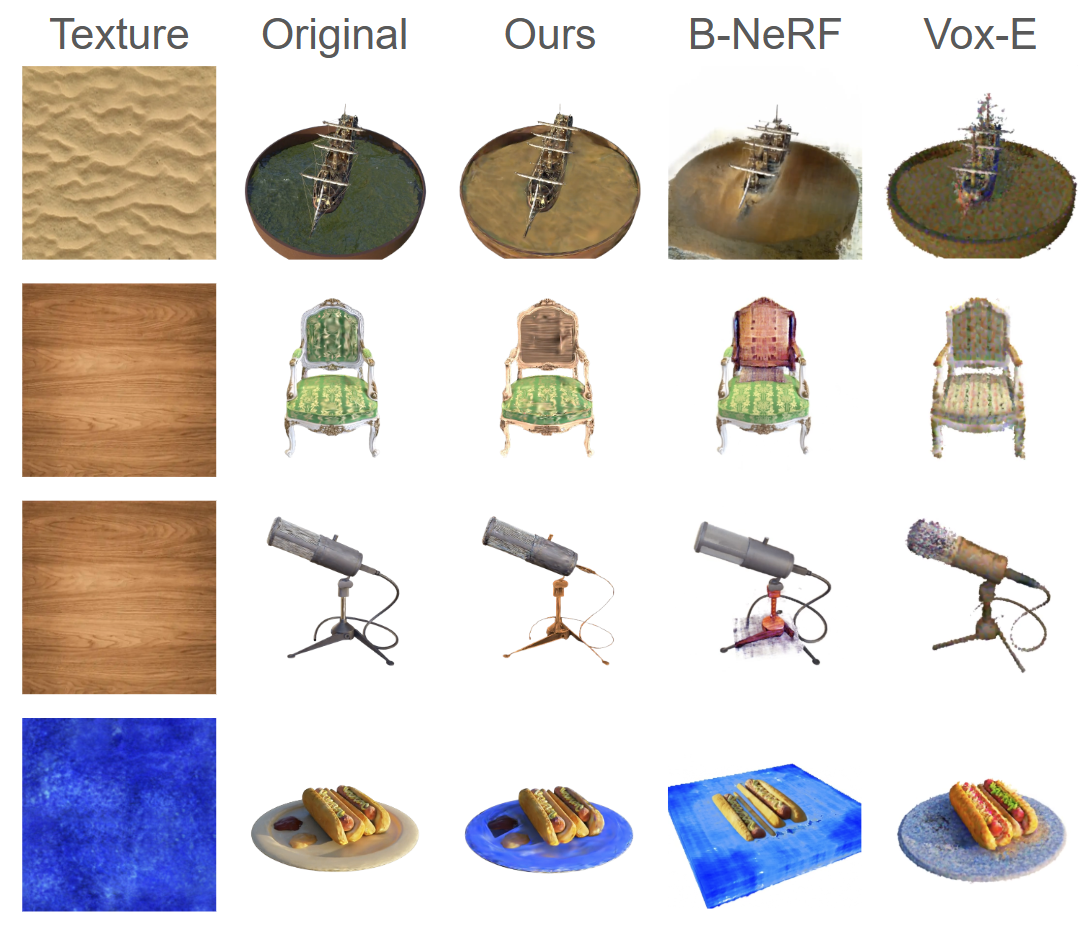}
    \captionof{figure}{Comparison of our method in local texture editing of ours and baselines.}
    \label{fig:tex_compare}
\end{center}
\textbf{In our method:} For the ship, we selected the mask that corresponds to the water and indicate that the sand texture should be applied there. For the chair, we select the back and armrests. For the mic we select the stand and for the hotdog we select the plate.\\
\textbf{In BlendedNeRF:} To edit, a 3D box region and a corresponding prompt are required. For the ship, the box covers the xy plane, extending upwards to the ship's start, with the prompt 'dunes of sand'. The chair's box is above the seat cushion, including the armrests, using 'a wooden chair'. The mic's stand is boxed with 'a wooden stand'. For the hotdog, the box spans the xy plane, extending vertically to the hotdogs' start.\\
\textbf{In Vox-E:} We should specify our prompt as what we want the final image to be, as this is the input to the diffusion guidance. For the ship we use ‘a ship in sand’. For the chair we use ‘a chair with a wooden back’. For the mic we use ‘a microphone with a wooden stand’, and for the hotdog we use ‘a hotdog on a blue granite plate’.
\subsubsection{Analysis}
Vox-E allows users to use text prompts for editing object voxel grids but has notable limitations. The text prompt can't focus edits on specific areas, leading to unwanted changes, such as unnecessary coloring in the chair and microphone examples as in Figure~\ref{fig:tex_compare}. It also struggles with texture representation, producing rough, pixelated textures that don't match the intended edits, as seen in the plate and ship examples. Additionally, while Vox-E can change shapes, this sometimes results in unintended alterations.

BlendedNeRF can produce high-quality visuals but suffers from unintended artifacts and shape distortions due to its edit region being box-shaped, making precise edits difficult in intertwined areas. This issue is evident in examples like the ship, where sand spills out improperly, the chair with misplaced wooden panels, a mic with fuzzy artifacts, and a hotdog plate turned square. Unlike box-based edits, our mask-based approach allows for more precise region modifications. Additionally, BlendedNeRF struggles with texture definition, failing to produce detailed contours in sand or realistic wood grain, as highlighted in the ship and wood examples.
\begin{center}
    \centering
    \captionsetup{type=figure}
    \includegraphics[width=.45\textwidth]{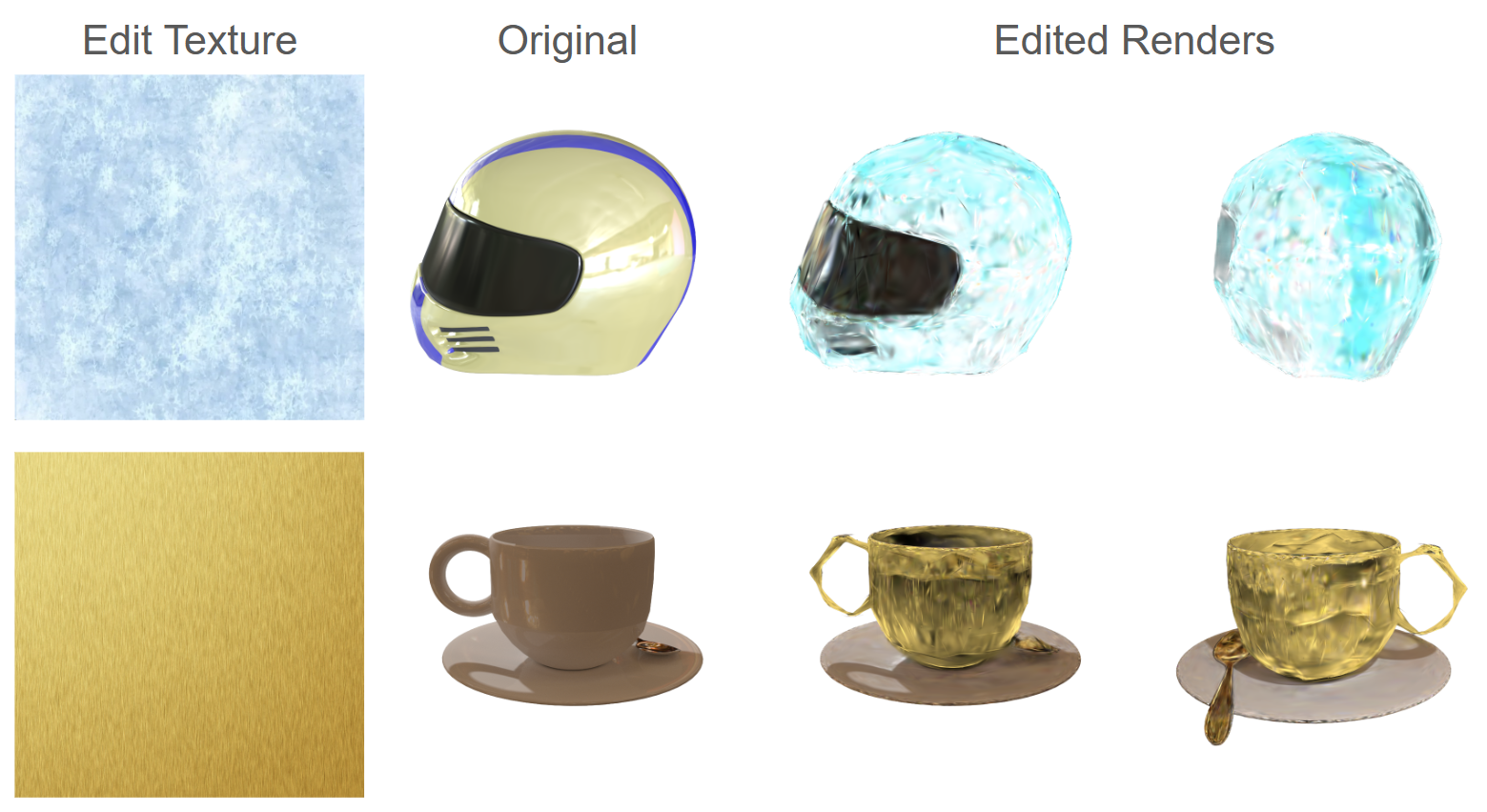}
    \captionof{figure}{Making an ice helmet and a gold foil coffee cup from RefNeRF.}
    \label{fig:helmet_and_coffee}
\end{center}
\begin{center}
    \centering
    \captionsetup{type=figure}
    \includegraphics[width=.45\textwidth]{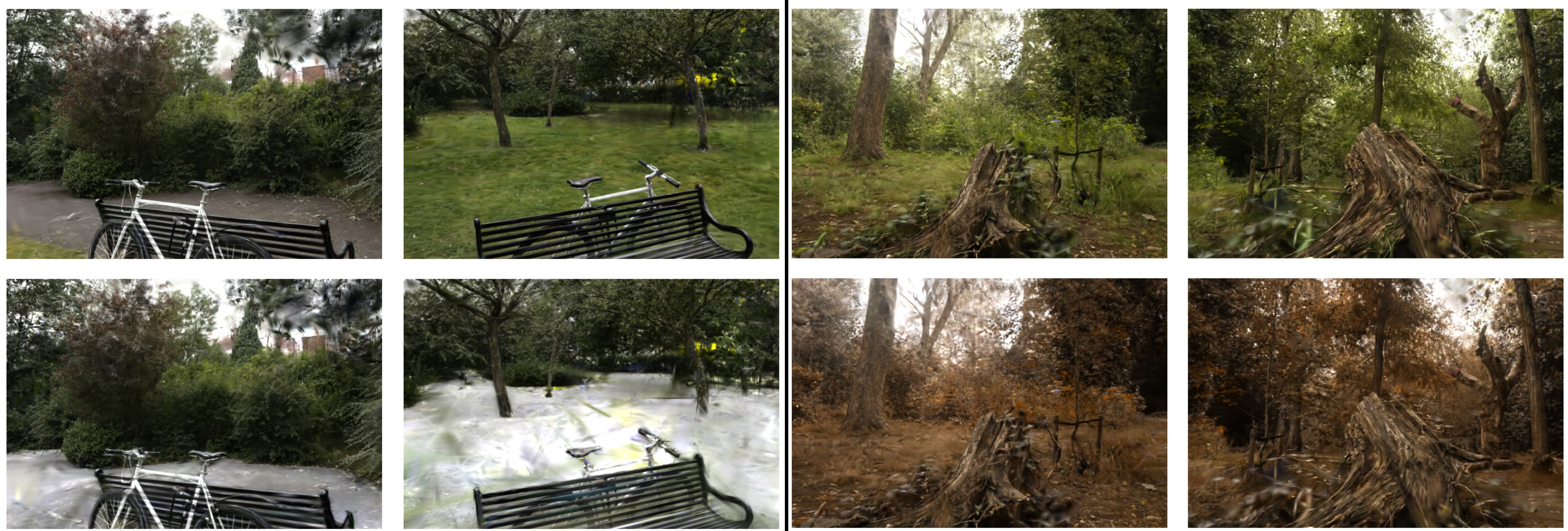}
    \captionof{figure}{Applying a snow and ice texture to the scene (left), and turning the stump scene to fall (right)}
    \label{fig:garden}
\end{center}
\subsection{Color Editing}
\label{sec:col_editing_results}
\begin{center}
    \centering
    \captionsetup{type=figure}
    \includegraphics[width=.45\textwidth]{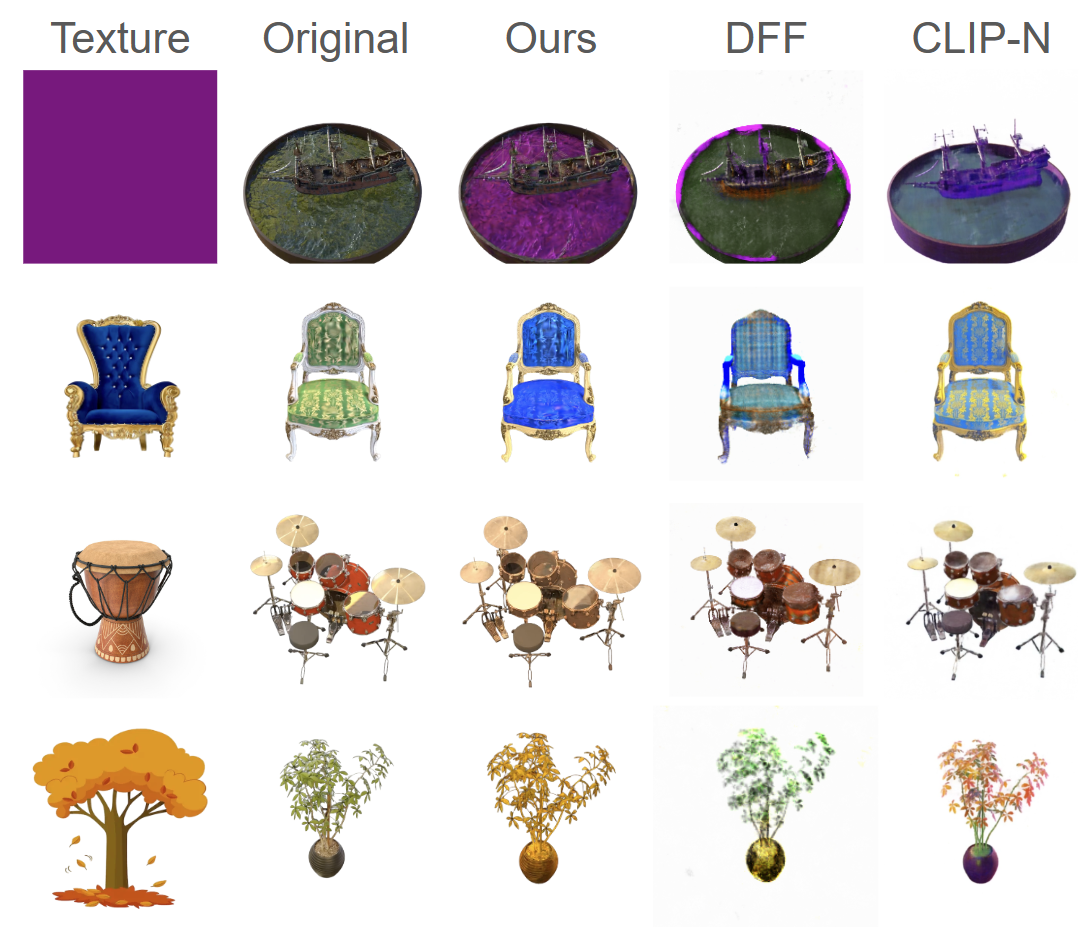}
    \captionof{figure}{Showing local and correspondance based color editing across ours and baselines.}
    \label{fig:col_compare}
\end{center}
\textbf{In our method:} The purple color is applied to the water region of the image for the ship. For the chair, drums, and plant, appropriate color regions are automatically extracted with mask matching and applied onto the dataset views.\\
\textbf{In D.F. Fields:} This method uses a text phrase for the new color and a filter phrase for object selection, where simply naming the object works best due to D.F. Fields' difficulty in selecting object subparts. Adding 'background' to the filter phrase improves performance. Color prompts are specific, like 'purple water' for the ship, 'brown drums', 'golden plant', and 'blue and gold chair'. Excessively detailed prompts tend to reduce effectiveness.\\
\textbf{In CLIP-NeRF:} In CLIP-NeRF, we specify a prompt as a sentence of what we want to see in the result sentence. For the ship, this is ‘a ship in purple water’. For the chair, it is ‘a chair with blue cushions and a gold frame’. For the drums is ‘brown drumset with bronze cymbols’, and for the ficus it is ‘plant with fall-colored leaves’.
\subsubsection{Analysis}
DF Fields effectively changes colors across broad areas but struggles with precision in smaller regions. This model cannot finely detail or recolor small parts of images, exemplified by unchanged colors in the instruments~\ref{fig:col_compare} and difficulty in differentiating the ship from its surrounding water. Additionally, trying to specify exact RGB colors through text often leads to discrepancies between intended and actual colors.

CLIP-NeRF produces attractive results but can deviate from precise prompts. For instance, a request for a ship in purple water resulted in both the ship and a bucket being colored purple instead of just the water. This indicates a challenge with the precision of text-based editing compared to direct mask and color adjustments. Other examples include unintended additions like a gold pattern on a chair meant to be simple, minimal changes to drum colors, and an unexpectedly colored pot on a plant, showcasing the limitations in accurately reflecting user intentions through text/vector embedding.

\subsection{Inherited Limitations}
We inherit a few limitations from pretrained components utilized in our method. SAM, which performs segmentation on selected views can sometimes fail to generate fine grained masks from certain angles, lumping together two parts of an object. When this happens, an edit that would have normally been constrained to one area in that particular view can sometimes bleed into other areas. This is rare in most scenes, but can occur in complicated scenes, or object sections with ill-defined boundaries. Also, using SAM to select masked regions means that our method cannot perform edits to the 3D geometry of the object.\\
The NNFMloss function is great at copying texture and overall style from a source area to destination area, but makes the result unreflective. This is seen in Figure~\ref{fig:helmet_and_coffee}, where the original surface of objects was reflective, and applying a new texture unintentionally overwrote those effects. Likewise, if the edit image’s texture contains any such light scattering effects, these are not carried over onto the Gaussian Splat.
\subsection{Data}
We use publically available synthetic datasets from NeRF~\cite{mildenhall2020nerf} and RefNeRF~\cite{verbin2022refnerf}, as well as real scenes from Mip-NeRF~\cite{barron2022mipnerf360} and NeRDS360~\cite{irshad2023neo360}. We use internet images under cc-license for the style conditioning.
\subsection{Computation Time}
We find that our method performs much faster when implemented on Gaussian Splats, showing that color and style losses can be applied faster on this representation. In our experiments, running our method on top of standard NeRFs took more iterations to transfer style, and each iteration also ran slower, showing that it is easier to change color and texture on a Gaussian Splat. We include timings for the other baselines we tested in Table~\ref{timing}, along with the timing for SINE from that paper.
\begin{center}
\captionsetup{type=table}
\begin{tabular}{c  c c c c} 
\hline
  & Avg Time (Mins)\\ [0.5ex] 
 \hline
 Vox-E & 52\\
 DF Fields & 33\\
 CLIP-NeRF & 35\\
 BlendedNeRF & 118\\
 SINE & 720\\
 \hline
 Ours (NeRF) & 40\\
 \textbf{Ours (GS)} & \textbf{21}\\
 \hline
\end{tabular}
\captionof{table}{Average runtimes we observed for obtaining quality results for each method on a single NVIDIA A40 GPU.}
\label{timing}
\end{center}
\subsection{User Study}
In the user study, we seek to understand how users perceive our method as compared with leading baselines. Since the text prompts we chose for each of these baselines detailed in Sections \ref{sec:tex_editing_results} and \ref{sec:col_editing_results} are a faithful representation of the edit we intend to express with the conditional image we use for our method, we can compare against these baselines accurately. We solicit feedback on the user preferences from 38 people, and asked about their expertise with generative models. The ratings were requested via a Google Form. \textit{Ten were familiar with generative computer vision and twenty-eight were not.}
\subsubsection{Texture}
As displayed in Figure~\ref{fig:tex_compare} in the paper, we test on the baselines of Vox-E and BlendedNeRF. For each of the texture editing instructions, we ask the user to choose the result that best transfers the texture shown onto the specified area of the image, and specify the following instructions:
\begin{itemize}[leftmargin=.25in]
    \item Turning the water into sand
    \item Turning the chair back and frame into wood
    \item Turning the plate blue granite
    \item Turn the mic stand wood
\end{itemize}
\begin{center}
\captionsetup{type=table}
\begin{tabular}{c  c c c} 
\hline
 Object & Ours & BlendedNeRF & Vox-E\\ [0.5ex] 
 \hline
 Ship & \textbf{86.8\%} & 7.9\% & 5.3\%\\
 Chair & \textbf{63.2\%} & 26.3\% & 10.5\%\\
 Mic & \textbf{68.4\%} & 13.2\% & 18.4\%\\
 Plate & \textbf{73.7\%} & 5.3\% & 21.1\%\\
 \hline
\end{tabular}
\captionof{table}{Percent of users who preferred each method for texture editing.}
\label{timing}
\end{center}
In all cases, our method was favored by most users. For the ship example, it received high preference due to BlendedNeRF's sand spilling out of bounds and Vox-E's unrepresentative grainy texture. In the chair scenario, 63.2\% preferred our method, noting it provided a reasonable texture, whereas BlendedNeRF was a close second. Vox-E’s inaccuracies, such as miscoloring parts of the mic, were noted by attentive users. Our plate design also won majority preference, with BlendedNeRF's version turning square and Vox-E erasing condiments. Similarly, our method was the top choice for the mic, as BlendedNeRF's edits introduced unwanted artifacts.

\subsubsection{Color}
Here we test against Distilled Feature Fields and CLIP-NeRF, with three global style transfer examples from conditional images, and one local color editing example as in Figure~\ref{fig:col_compare}. For the global color transfer, we explain the concept of correspondence in simple English, by asking the user to select the result which takes on the color scheme of the edit image applied onto the original. For the local color transfer on the ship example, we mention that the goal is to turn the water in the image purple.
\begin{center}
\captionsetup{type=table}
\begin{tabular}{c  c c c} 
\hline
 Object & Ours & DF Fields & CLIP-NeRF\\ [0.5ex] 
 \hline
 Ship & \textbf{84.2\%} & 7.9\% & 7.9\%\\
 Chair & \textbf{73.7\%} & 5.3\% & 21.1\%\\
 Drums & \textbf{73.7\%} & 10.5\% & 15.8\%\\
 Plant & \textbf{65.8\%} & 2.6\% & 31.6\%\\
 \hline
\end{tabular}
\captionof{table}{Percent of users who preferred each method for color editing.}
\label{timing}
\end{center}
In color editing, our method again won over 60\% of user preference in each scenario. For the ship, our recolor was favored as Distilled Feature Fields partially recolored the tray border and CLIP-NeRF mistakenly colored the ship, not the water. Our chair was preferred for its accurate gold frame and blue cushion, matching the throne, though CLIP-NeRF also attracted 21.1\% of users with its intriguing, albeit unintended, result. Both DF Fields and CLIP-NeRF struggled with coloring the drums correctly, leading to low preference. For the plant, DF Fields failed to alter its green color, while CLIP-NeRF’s reddish fall colors caught some interest, but overall, our method was seen as most accurately reflecting the intended edits.

\section{Conclusion}
In this work, we have introduced a robust and flexible method for editing color and texture of 3D images and scenes. We provide interfaces to copy style from an edit image or manually specify changes, enabling creative appearance editing for a variety of applications. Our key innovation, DINO-based mask matching, runs quickly and contains edits to discrete regions, leading to higher quality than other methods. Future work could explore how to make 3D consistent shape changes to these discrete regions in addition to color and texture, without compromising on resulting 3D scene quality like most other current methods do. Overall, we showcase our method’s unique input expressivity and resulting 3D model quality on a variety of objects and scenes, proving it is well suited for creative applications.
{
    \small
    \bibliographystyle{ieeenat_fullname}
    \bibliography{main}
}
\section{Appendix}
\subsection{User Workflow}
\begin{center}
    \centering
    \captionsetup{type=figure}
    \includegraphics[width=.45\textwidth]{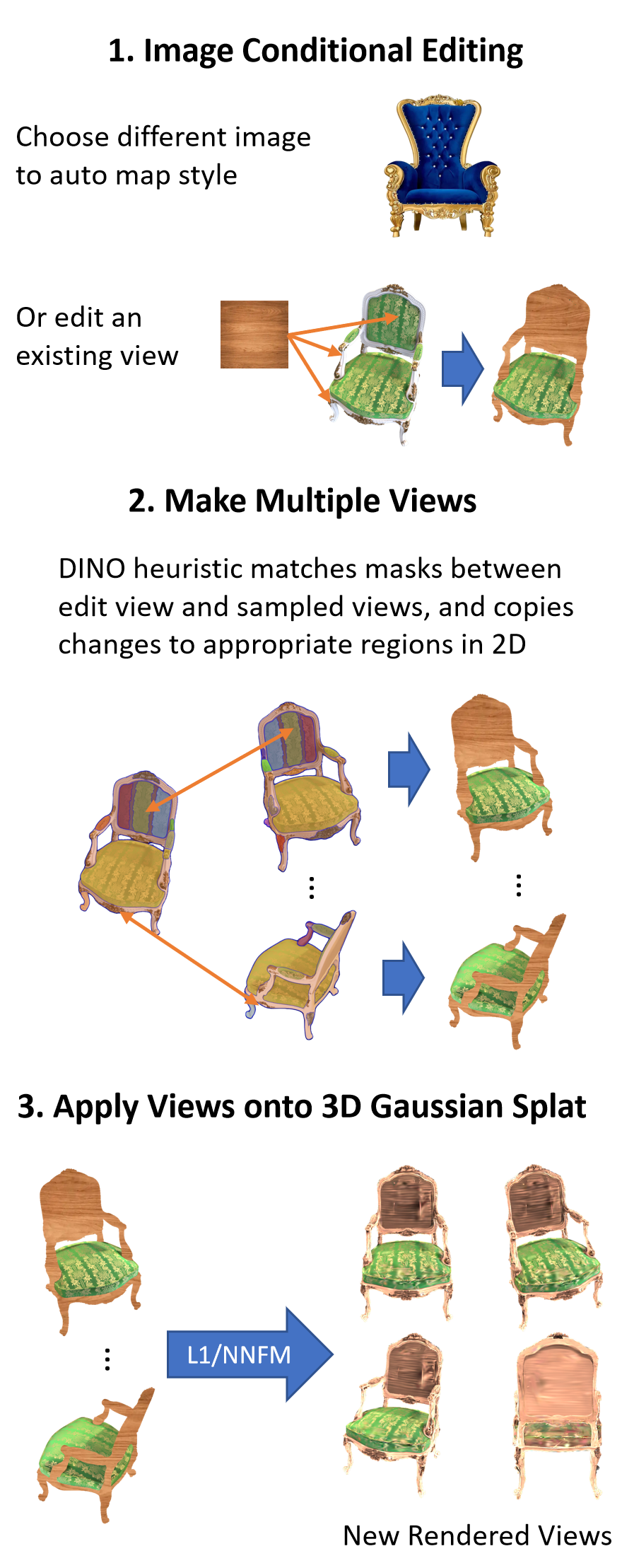}
    \captionof{figure}{User initially chooses an image to automatically extract correspondences from, or makes an edit to an existing view.}
    \label{fig:multi_bonsai}
\end{center}
\subsection{Additional Editing Results}
\begin{center}
    \centering
    \captionsetup{type=figure}
    \includegraphics[width=.5\textwidth]{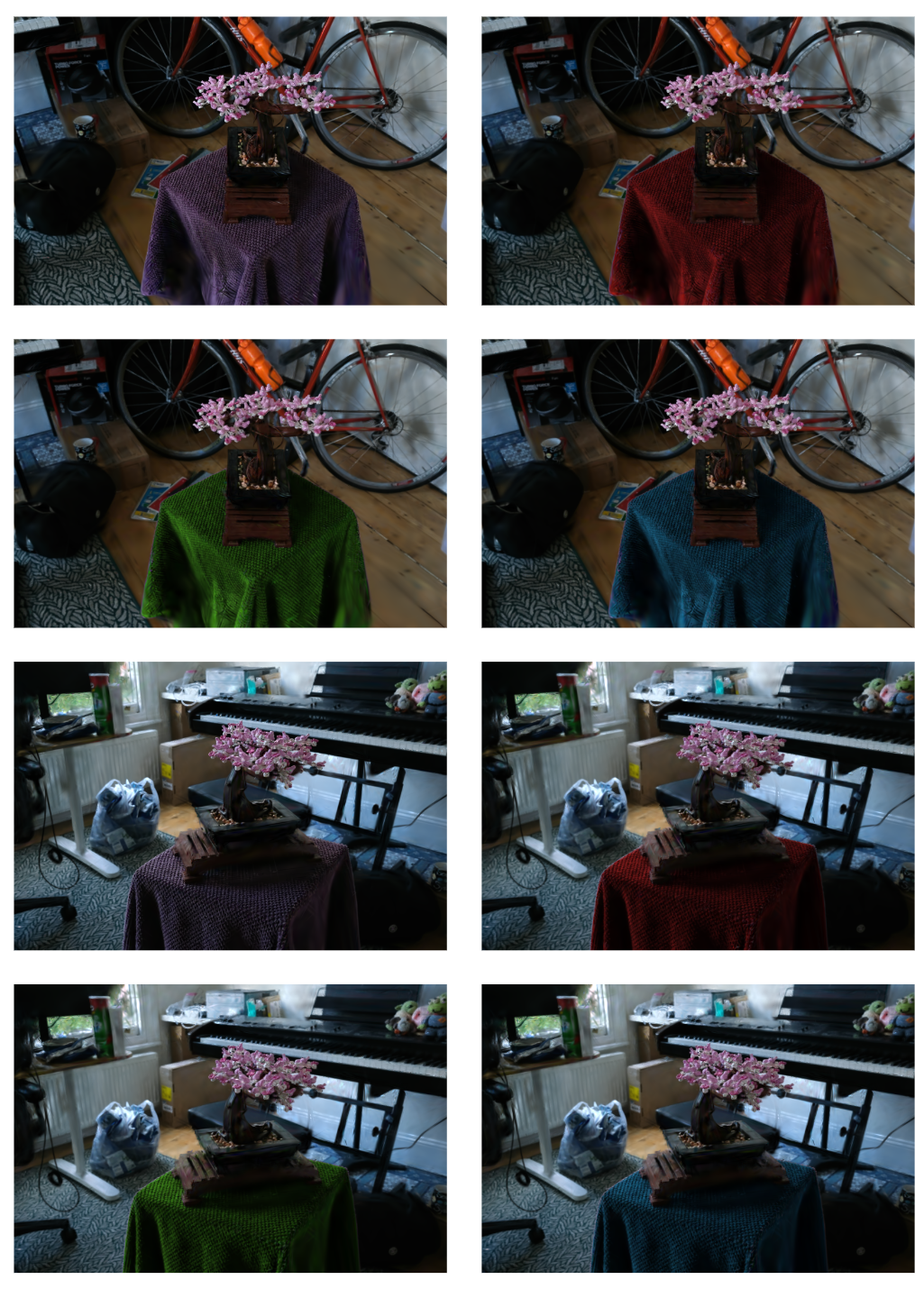}
    \captionof{figure}{Our method is able to perform bounded and accurate color changes across views, even in cases of a cluttered background with numerous object masks.}
    \label{fig:multi_bonsai}
\end{center}
\begin{center}
    \centering
    \captionsetup{type=figure}
    \includegraphics[width=.4\textwidth]{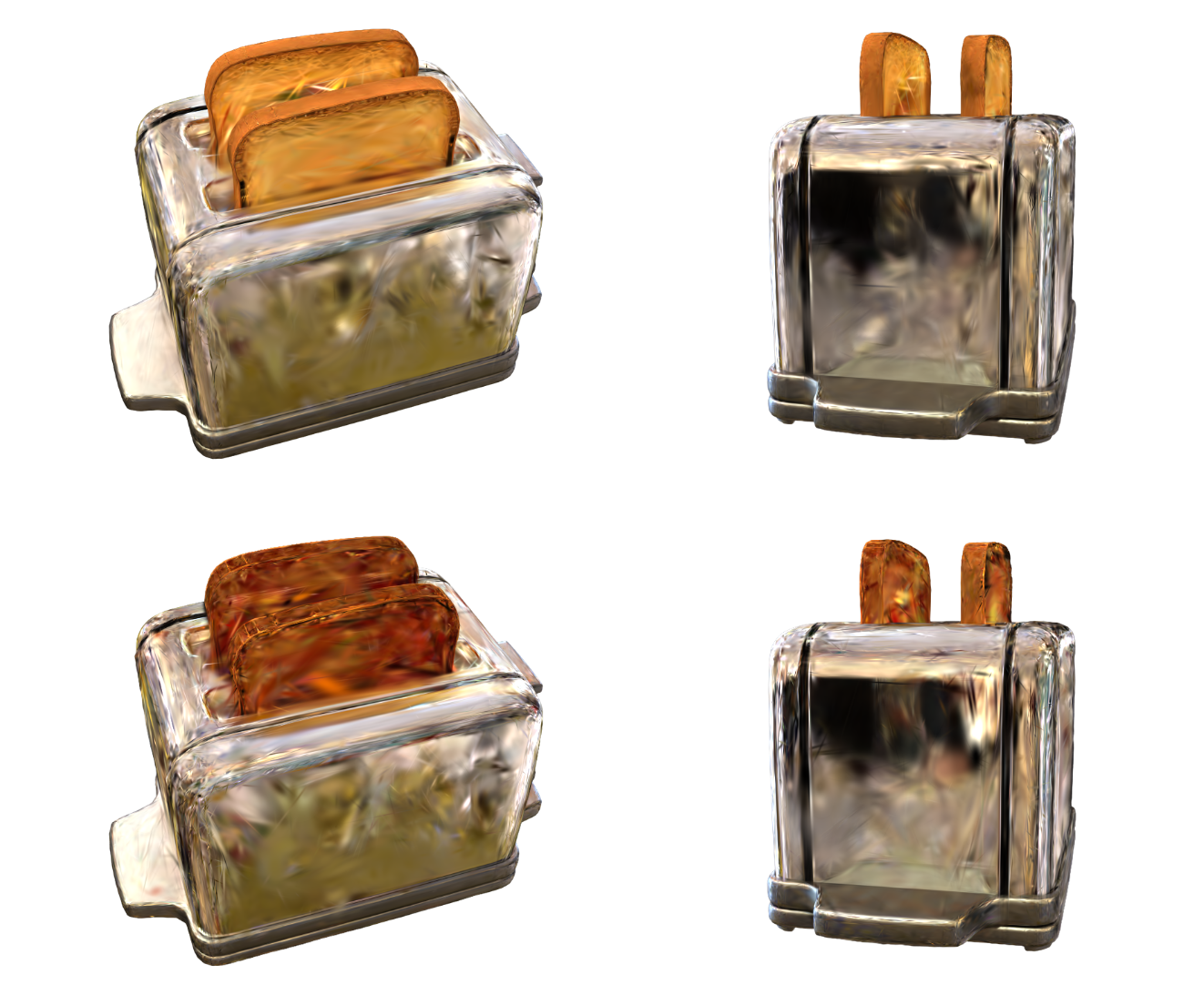}
    \captionof{figure}{Toasting the toast without affecting toaster.}
    \label{fig:garden}
\end{center}
\begin{center}
    \centering
    \captionsetup{type=figure}
    \includegraphics[width=.5\textwidth]{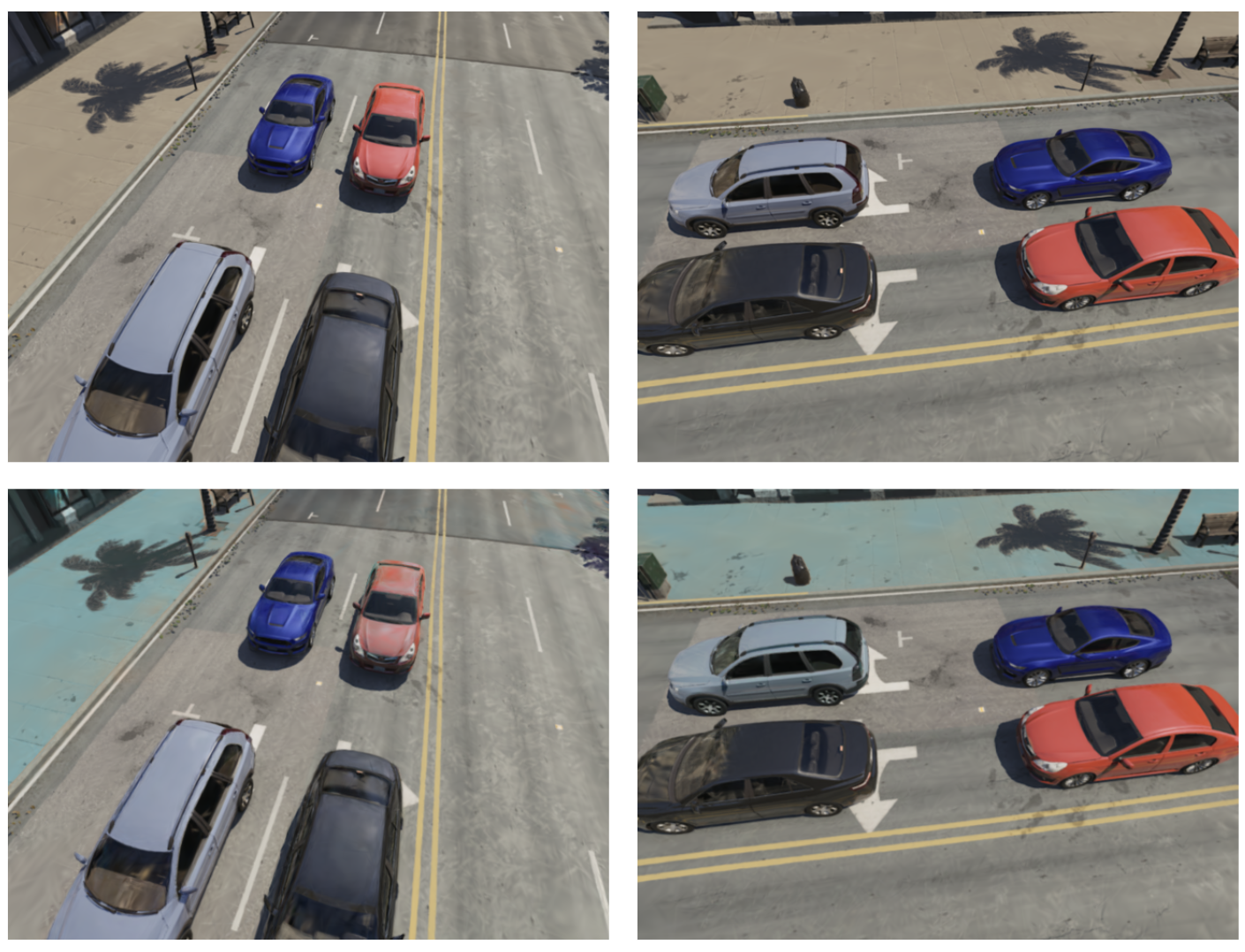}
    \captionof{figure}{Turning the sidewalk light blue without affecting the street.}
    \label{fig:sidewalk_light_blue}
\end{center}
\begin{center}
    \centering
    \captionsetup{type=figure}
    \includegraphics[width=.5\textwidth]{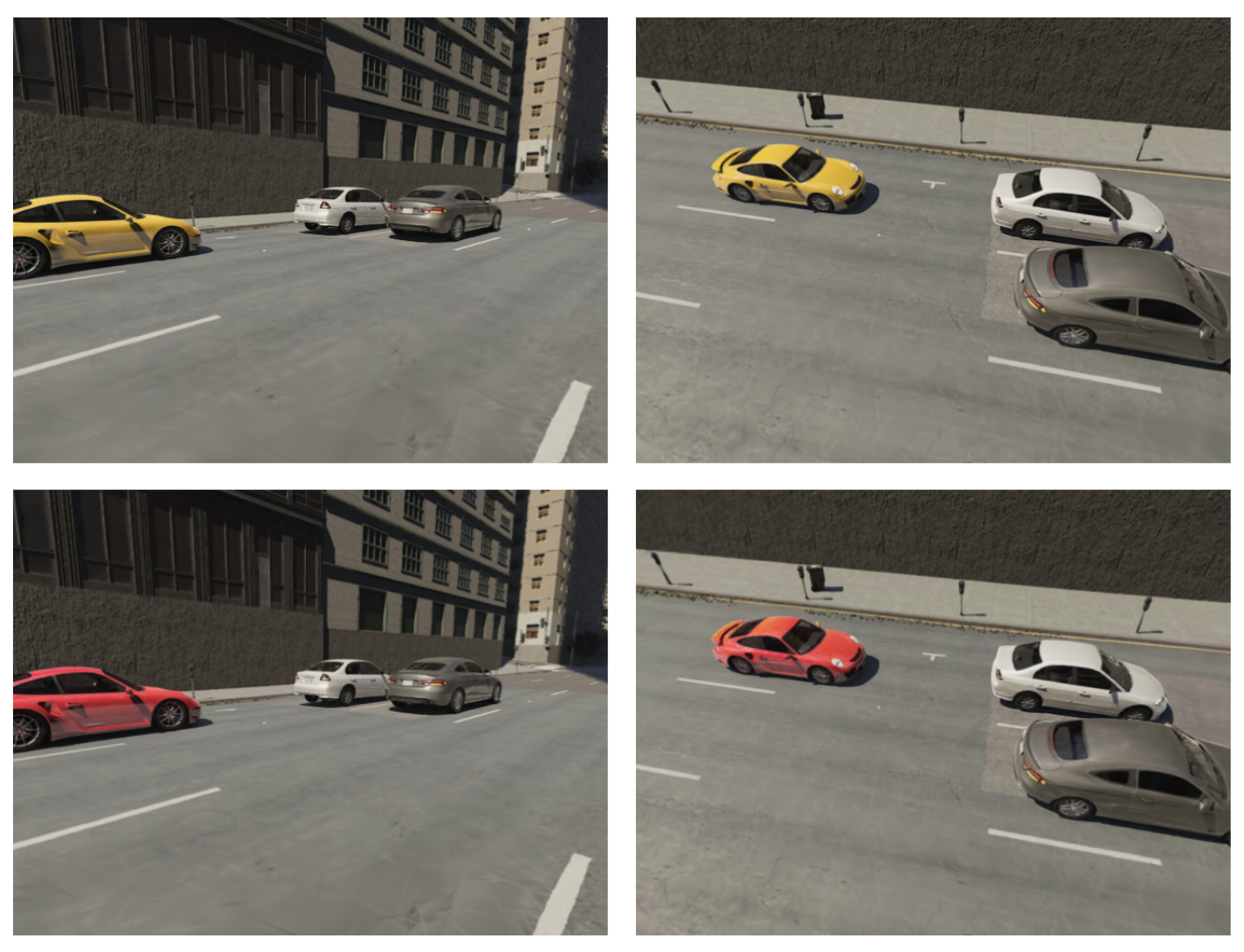}
    \captionof{figure}{Turning just the yellow car red.}
    \label{fig:grant_redcar}
\end{center}
\begin{center}
    \centering
    \captionsetup{type=figure}
    \includegraphics[width=.4\textwidth]{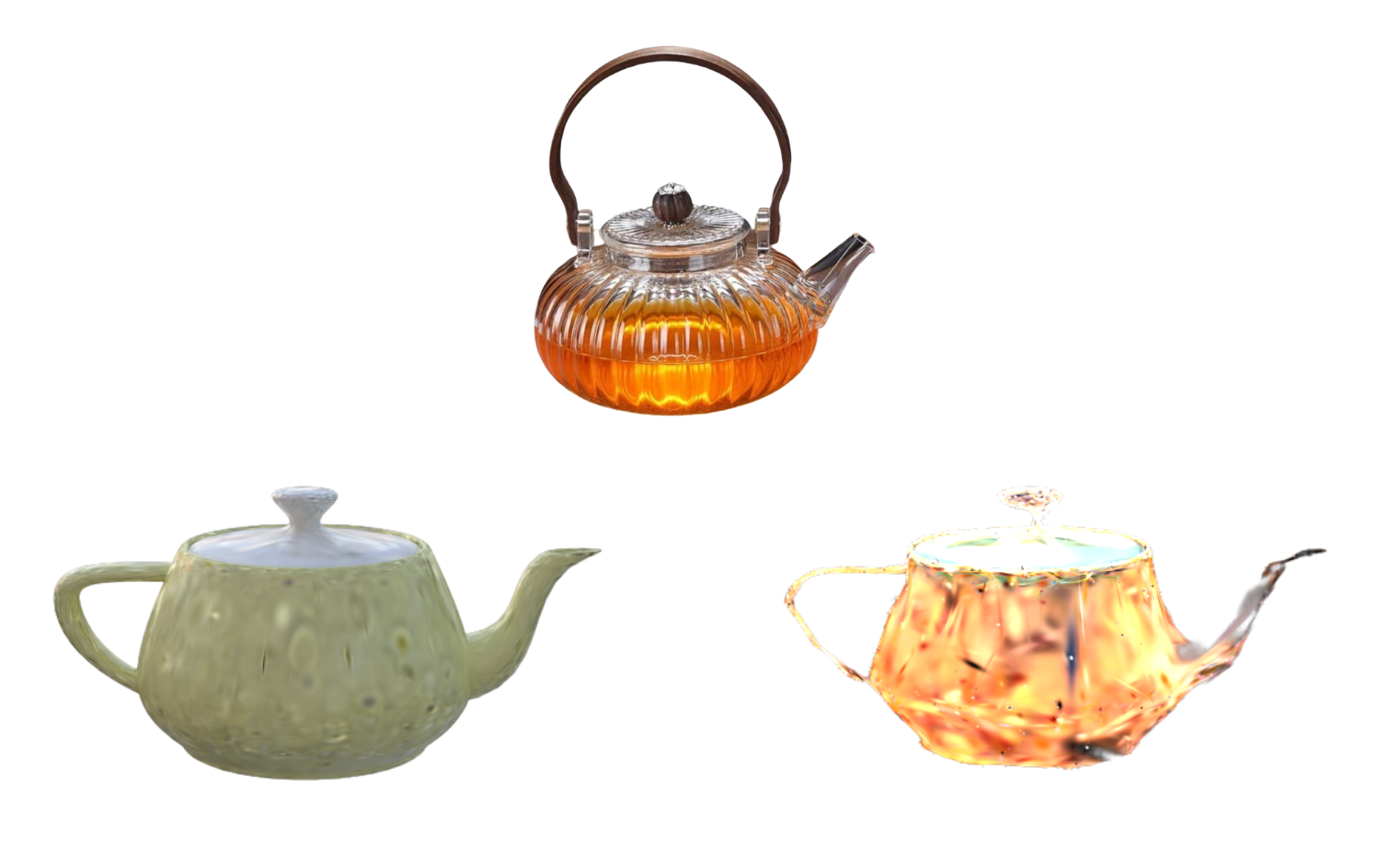}
    \captionof{figure}{One limitation of our method is that it struggles to accurately copy textures from layered objects like liquid behind a glass.}
    \label{fig:teapot}
\end{center}
%


\end{document}